\documentclass[a4, 10pt, conference]{ieeeconf}  

\IEEEoverridecommandlockouts                              

\usepackage[dvipdfmx]{graphicx} 
\usepackage{times}
\usepackage[frozencache,cachedir=.]{minted}
\usepackage{url}
\usepackage{CJKutf8}

\title{\LARGE \bf
Tourist Guidance Robot Based on HyperCLOVA
}

\author{Takato Yamazaki$^{1}$, Katsumasa Yoshikawa$^{1}$, Toshiki Kawamoto$^{1,2}$, Masaya Ohagi$^{1}$,\\Tomoya Mizumoto$^{1}$, Shuta Ichimura$^{1}$, Yusuke Kida$^{1}$, Toshinori Sato$^{1}$
\thanks{$^{1}$These authors are with LINE Corporation, Tokyo, Japan. Representative email address: {\tt\small takato.yamazaki@linecorp.com}}%
\thanks{$^{2}$Toshiki Kawamoto also belongs to Tokyo Institute of Technology, Tokyo, Japan}%
}

\begin{document}

\maketitle
\thispagestyle{empty}
\pagestyle{empty}

\begin{abstract}
This paper describes our system submitted to Dialogue Robot Competition 2022.
Our proposed system is a combined model of rule-based and generation-based dialog systems.
The system utilizes HyperCLOVA, a Japanese foundation model, not only to generate responses but also summarization, search information, etc.
We also used our original speech recognition system, which was fine-tuned for this dialog task.
As a result, our system ranked second in the preliminary round and moved on to the finals.
\end{abstract}

\section{INTRODUCTION}
This paper describes our system submitted to Dialogue Robot Competition 2022.
The dialog task for this competition is to develop a tourist guidance system that utilizes a humanoid robot \cite{higashinaka-2022, minato-2022}.
In our system, we have developed a scenario-based dialog system using HyperCLOVA, a Transformer-based Japanese foundation language model with 82B parameters.
We also developed a speech recognition system, which we fine-tuned for this tourist guidance domain.
As a result of the preliminary round, our system ranked second and moved on to the final round.
The evaluation was performed through a survey after the dialog session.
The results show that our system could achieve the best trustworthiness score among all the submitted systems, which could be the advantage of flexible responses from HyperCLOVA.
However, our model has a low naturalness score compared to the other metrics, which may be due to our method that does not fixate utterances in a rule-based manner.
Overall, the scores on all the evaluation metrics are still far from the maximum score, which indicates that there is space for improvement in many aspects, such as response quality, response time, body movement, etc.

\section{COMPETITION DESCRIPTION}
The dialogue task in this competition simulates a situation in which the robot acting as a counter-sales person at a travel agency accommodates customers' requests.
The customer's objective is to decide on a single tourist sight to visit, by consulting with the robot to choose from two candidates.
The robot is pre-informed by the organizer's system which tourist sight to strongly recommend and is expected to persuade the customer while showing appropriate hospitality.
The designated conversation time is about five minutes.

During the competition, a booth simulating a travel agent's counter is set up in a fixed place.
The preliminary round was held in \textit{The National Museum of Emerging Science and Innovation} in Odaiba Tokyo, Japan.
In the booth, two chairs are facing each other and a desk is in between.
The humanoid robot sits in one chair and a customer, who randomly showed up at the booth, sits in the other.
In addition, there is a microphone on the desk and there is a camera behind the humanoid.

Contestant systems are allowed to use sensor information from the microphone and camera, and also have access to the tourist sight database which contains a summary, business hours, facility information, etc.
The contestants' system is expected to control the humanoid robot's body (e.g. viewpoint, pose, head inclination), expression, and speech.
The contest organizers provide intermediate software to allow contestants to easily control the humanoid.

The evaluation is based on the customer's survey, which is answered after the dialog session.
Detailed descriptions are in \cite{minato-2022}.

\begin{figure}[t]
    \centering
    \includegraphics[scale=0.8]{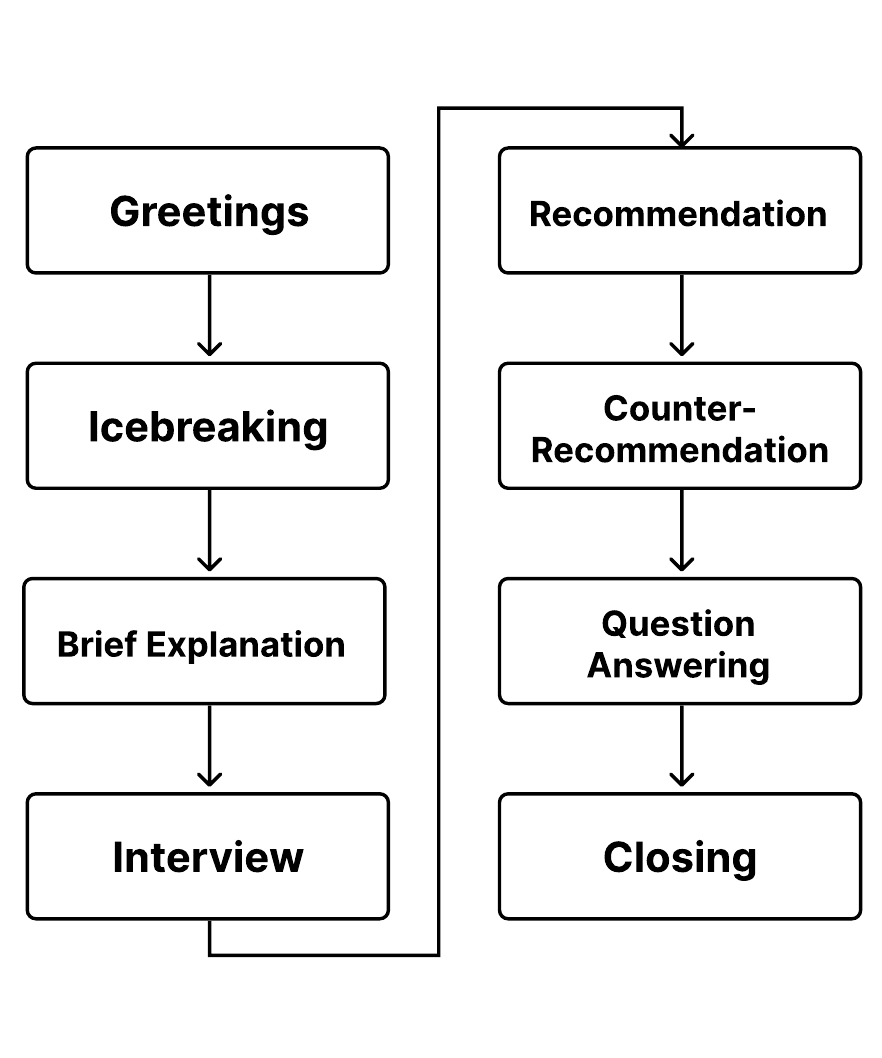}
    \caption{Dialog Scenario of Our System}
    \label{fig:scenario}
\end{figure}

\section{SYSTEM OVERVIEW}

The system proceeds with the dialog task with a predefined scenario, shown in Figure \ref{fig:scenario}.
The robot, which we named \textit{Shoko}, starts with simple greetings and then does some chit-chat as an icebreaker.
Then, it briefly explains the purpose of this conversation and the selected tourist sight.
After that, the robot asks some questions to obtain the customer's information.
Based on that result, the robot recommends and counter-recommends the tourist sights.
Finally, there will be a question-answering time and then finish with a greeting in the closing phase.

Our system is based on HyperCLOVA\cite{kim-etal-2021-changes}, which is a system containing a Japanese foundation language model with 82B parameters.
Previous works with HyperCLOVA showed that using few-shot learning (or prompt learning) enables generating fluent and interesting responses in both open-domain \cite{yamazaki2021} and situated \cite{kawamoto2021} dialog tasks.
In this competition, unlike the previous works, we attempted to combine HyperCLOVA with the rule-based dialog to achieve a controllable task-oriented dialog system.
In our system, HyperCLOVA is used not only in generating responses, but also generating summaries, finding recommendation points, searching for information, etc.
Details of the HyperCLOVA usage will be explained in section \ref{sec:dialog-system}.
Note that all the following prompt examples in the figures are translated into English, and they are originally Japanese.

Throughout the system, we used a speech recognition system developed for this tourist guidance domain, which will be described in section \ref{sec:speech_recog}.
For the robot's facial expression, we made the robot smile most of the time, but when the generated utterance contained an exclamation mark, we made the robot's eyes open widely and eyebrows go up to look like it is surprised.
For the robot's movement, the robot looks at the monitor during the \textit{Brief Explanation} and \textit{Recommendation}, to attract the customers' attention to the monitor.
We also made the robot nod its head at random intervals when the user is speaking.

\section{DIALOG SYSTEM}\label{sec:dialog-system}

\subsection{Greetings and Icebreaker}
The system starts the conversation with a fixed conventional Japanese greeting and self-introduction.
The robot is programmed to bow during greetings, which is a common thing to do in Japanese culture.
In the greetings, the robot asks the customer to speak loudly, since the speech recognition system is vulnerable to noisy and unclear voices.

After the greetings, the system moves on to the icebreaker phase, in which the robot asks open questions about the customers' work.
We included this phase because of two reasons: 1) to let customers feel relaxed by letting them speak about themselves, and 2) humans tend to speak to the robot in very short and rough answers such as ``Yes'' or ``OK'', and the robot needs to inform them that it is capable of understanding free-form and long utterances.

Our system's icebreaker phase is structured as three turn conversation.
For the first turn, the robot asks the customer with a fixed utterance, ``What do you do for a living?'', and wait for the customer's responsibility.
For the second turn, our system uses HyperCLOVA to respond to the answer, by inputting the prompt shown in Figure \ref{fig:icebreak-q}.
It is designed to ask a follow-up question to the customer's work.
For the third turn, our system again uses HyperCLOVA to respond with a generous comment, by using a similar prompt used in the second turn.

\begin{figure}
    \centering
    \begin{minted}[breaklines, fontsize=\scriptsize, frame=single, breaksymbol=\quad, breaksymbolindentnchars=2]{text}
[Shoko, who is always cheerful, asks questions about the client's occupation.]

###
A client is a man in his late 40s.

Shoko(question): Nice to meet you! I am a tourist guide. What kind of work do you usually do?
Customer:I am a manager in an IT company.
Shoko(question):That's wonderful! What do you think is the most important for your management work?
...
    \end{minted}
    \caption{A Shot example for Question Generation in Icebreaking}
    \label{fig:icebreak-q}
\end{figure}

\subsection{Brief Explanation}

After the icebreaking, the system briefly explains the objective of this conversation and the selected tourist sights.
It first says ``I heard that you are deciding between two tourist sites,'' then asks ``have you ever been to either one of those?''.
The system then uses a regular expression to detect if the answer is yes or no.
We prepared fixed responses for both cases.

Then, the system briefly explains the two tourist sights.
Since there were not any summaries in the provided database, we generated a one-line summary by utilizing HyperCLOVA.
We handcrafted a prompt for the summarization task, in which the input is a long explanation of the tourist sight and the output is a one-line summary.

\subsection{Interview}
For the interview section, we apply a strategy that combined a conventional rule-based approach with a modern text generation approach.
The former allows us to obtain several essential pieces of information from users, and the latter provides users with some pertinent and comfortable comments based on their answers.
We prepare a set of questions to find out user preferences, in advance.
The answers from users are always analyzed by a simple slot-filling function.

\begin{table*}[]
\centering
\caption{Prepared Questions for the Interview Phase}
\begin{tabular}{|c|l|l|l|l|}
\hline
ID & \multicolumn{1}{l|}{Type}  & Item                          & Question                                        & Expected Answers and Next ID                                                                       \\ \hline
1  & {Mandatory} & {Participants} & Who are you traveling with?                     & \begin{tabular}[l]{@{}l@{}}alone / friend (to 6)\\ family (to 2)\end{tabular} \\ \cline{1-1} \cline{4-5} 
2  &                            &                               & Do you plan to bring your children with you?    & \begin{tabular}[c]{@{}l@{}}Yes (to 3) \\ No (to 6)\end{tabular}               \\ \cline{1-1} \cline{4-5} 
3  &                            &                               & How old are your children?                      & \$N years old (to 6)                                                          \\ \cline{1-1} \cline{3-5} 
4  &                            & Transportation                & Are you using a car for this trip?              & Yes / No (to 5)                                                               \\ \cline{1-1} \cline{3-5} 
5  &                            & Points of Interest            & Is there anything you focus on when you travel? & Open answer (to next phase)                                                                      \\ \hline
6  & {Loc-wise}  & Point1                        & e.g. Do you like to have magical experiences?   & Yes / No (to 7)                                                               \\ \cline{1-1} \cline{3-5} 
7  &                            & Point2                        & e.g. Do you like detective dramas?              & Yes / No (to 8)                                                               \\ \cline{1-1} \cline{3-5} 
8  &                            & Point3                        & e.g. Do you like to play by the water?          & Yes / No (to 4)                                                               \\ \hline
\end{tabular}
\label{tab:questions}
\end{table*}

We list the interview questions in Table \ref{tab:questions}.
There are two types of questions, the \emph{mandatory questions} which our system always asks, and the \emph{location-wise} questions which our system asks by user-selected location.
First, we mention two mandatory questions. They are fixed and selected according to the solid if-then rules.

\paragraph{Participants}
Our system always asks the first question, ``Who are you traveling with?''
This question is for recommending tourist sights based on the user's companions. 
For instance, if a user answers that he is traveling with his family, our system recommends him to a family-friendly location.
The first question branches into multiple due to the user's answers listed in Table \ref{tab:questions}

\paragraph{Transportation}
Another mandatory question is ``Are you using a car for this trip?''
Since some locations have no parking space and some are far from any train stations, it is desirable to be able to provide local access information.

\paragraph{Points of Interest}
In the last part of our interview section, our system asks ``Can you tell us what points are important to you to enjoy your travel?''
Though we do not exploit the answer to this question for our recommendation, that gives the user a sense of satisfaction by listening to itself.

\paragraph{Question Generation}
Our \emph{location-wise} questions are generated by HyperCLOVA for each sight location.
Figure \ref{fig:question_generation_prompt} shows our specific prompt which converts a sight summary into questions.

\begin{figure}[t]
    \centering
    \begin{minted}[breaklines, fontsize=\scriptsize, frame=single, breaksymbol=\quad, breaksymbolindentnchars=2]{text}
[Generate questions from a summary of sightseeing location.]

[Summary]
Daiba Park
The site of a gun battery built by the Edo Shogunate in preparation for foreign ships after Perry's arrival in Japan...

[Questions]
- Do you like the history of the Edo period?
- Do you like novels depicting the end of the Edo period?
    \end{minted}
    \caption{Prompt for Question Generation}
    \label{fig:question_generation_prompt}
\end{figure}

All questions are related to the sight and always formatted ``Do you like --- ?''
For example, the summary of \emph{Tokyo Trick Art Museum} has ``It is a place where you can have a magical experience by optical illusions.''
In this case, HyperCLOVA may generate ``Do you like to have magical experiences?''
For each sight, HyperCLOVA generates 10 questions and our system selects at most three unique questions. 
Because the question generation process is often time-consuming, we handled this process in advance.
All location-wise questions are corresponding to the recommendations mentioned in Section \ref{sec:recommendation}.

\paragraph{Comment Generation}
In addition to asking questions, our system responds to users' answers and gives them appropriate comments.
When commenting, our system repeats what the user has said with the nodding motion.
We completely separate the question and comment generation process.
We build another prompt shown in Figure \ref{fig:comment_generation_prompt} which generates only comments without any questions.
If this prompt accidentally generates a comment including a question, our system filters it out and re-generates the other comment.

\begin{figure}
    \centering
    \begin{minted}[breaklines, fontsize=\scriptsize, frame=single, breaksymbol=\quad, breaksymbolindentnchars=2]{text}
[Shoko will give a polite phase. She repeats what the customer has said.]

Shoko: We would love for your children to enjoy the trip. May I ask how old your children are?
User: They are 5 and 2 years old.
Shoko: 5 and 2 years old, Okay. I would suggest a place where children can play as well.
    \end{minted}
    \caption{Prompt for Comment Generation}
    \label{fig:comment_generation_prompt}
\end{figure}

\subsection{Recommendation and Counter-Recommendation}
\label{sec:recommendation}
For the recommendation phase, our system starts with a fixed utterance saying ``according to your information, we recommend \texttt{<sight name>}.''
The system then explains the appealing point of the recommended tourist sight.
Then, it explains why it is especially recommended to the customer, based on the information obtained in the previous \textit{interview} phase and a search system.
Finally, the robot does a counter-recommendation, which is to explain why the customer should not go to the other sight.

\paragraph{Search System}
During the recommendation, the system utilizes a search system to generate a convincing and factual response.
First, as an additional data resource to answer the question, we preliminarily crawled two tourist information websites, Jalan\footnote{\url{https://www.jalan.net}} and TripAdvisor\footnote{\url{https://www.tripadvisor.jp}}. The crawled data includes basic information such as business hours and transportation access, as well as reviews by users and their review score. The crawled data was formatted into a database using Elasticsearch\footnote{\url{https://www.elastic.co}}.

While this resource is useful to answer the question, if all information of each site is used as a prompt for QA, the information will be scattered and the prompts will be unnecessarily long. Therefore, we utilized HyperCLOVA to extract only relevant basic information from the speaker's questions.
As a prompt, we first write the instruction ``Extract only relevant information from the following information,'' and then listed the basic information of a tourist site. And by describing multiple samples that were prepared as templates, combining assumed queries and basic information, we performed few-shot information extraction. The example is shown in Figure \ref{fig:information_extraction_prompt}.

\begin{figure}
    \centering
    \begin{minted}[breaklines, fontsize=\scriptsize, frame=single, breaksymbol=\quad, breaksymbolindentnchars=2]{text}
Extract only what is relevant from the following information.

Business hours: Opening hours: 11:00 - 21:00 (final admission 20:30) Closed: Various days
Location: DECKS Tokyo Beach Island Mall 4F, 1-6-1 Daiba, Minato-ku, Tokyo 135-0091, Japan MAP
Access: Train: 2-minute walk from Odaiba Kaihin Koen Station on the Yurikamome Line, a 5-minute walk from Tokyo Teleport Station on the Rinkai Line
Charge: Adult: 1,000yen (High school students and above) Child: 700yen (4 years old - junior high school student)

Question: How much is it?
Answer: Charge: Adult: 1,000yen (High school students and above) Child: 700yen (4 years old - junior high school student)

Question: Where is it?
Answer: DECKS Tokyo Beach Island Mall 4F, 1-6-1 Daiba, Minato-ku, Tokyo 135-0091, Japan MAP

Question: How much does this place cost?
    \end{minted}
    \caption{Prompt for Information Extraction}
    \label{fig:information_extraction_prompt}
\end{figure}

\paragraph{Recommendation}
The recommendation starts with explaining why the sight is appealing, and this text is generated by HyperCLOVA.
The prompt use the summary of the sight and HyperCLOVA starts generating after ``This place is appealing because ''.

For the robot to be more convincing, the model explains why it is recommended especially to the customer.
To generate such a recommendation sentence, the system takes two steps: 1) create a short recommending point and 2) generate the response using the recommending point.

For step 1, The recommending point is generated from the answers of the interview phase.
If the customer says yes to any of the questions, we use the corresponding preprocessed recommendation point.
The preprocess is a translation task that takes a question as an input (e.g. "Will you be touring with a baby?") and a recommendation point as an output (e.g. ``\texttt{<sight>} is recommended to a family with babies.''), which is done with HyperCLOVA.
However, in some cases, there may be no suitable feature to explain to the customer.
Therefore, we also prepared a customer-independent recommendation point.
We extracted several features such as price, indoor/outdoor, distance from the station, number of reviews, etc. from the database, and find advantages that are independent of the customer's answer.
We take at most two recommendation points to pass on to step 2, in which the customer-independent recommendation point has lower priority than the customer-dependent ones.

For step 2, we define this response generation task as follows: $\{s, d, p\} \mapsto u$ where $s$ is the summary of the recommendation sight, $d$ is the searched data, $p$ is the recommendation point, and $u$ is the robot's utterance.
$d$ is obtained using the search system, by using the recommendation point from step 1 as the query.
We used HyperCLOVA to solve this task.

\paragraph{Counter-Recommendation}

To discourage the customer from going to the other sight, the system does a counter-recommendation, which explains why they should not go to that place.
To generate the final utterance, the system takes a similar two-step procedure as the recommendation explained above.
One example utterance is as follows: ``The Water Science Museum is also a good place to learn about science, but it has few reviews and is not very popular, so Madame Tussauds Tokyo is probably a better choice.''

\subsection{Question Answering}
One of the important roles of a counter-sales is to flexibly answer customers' open questions.
The robot first asks the customer if they have any questions.
If they do not have any questions, it skips to the closing phase, but if they do, the system generates the answer response by using HyperCLOVA.

We define this response generation task as follows: $\{s_1, s_2, d_1, d_2, r, q\} \mapsto a$, where $s_1, s_2$ are summary of the tourist sights, $d_1, d_2$ is the searched data (using the same search system in section \ref{sec:recommendation}) for each tourist sights, $r$ is the recommended sight, $q$ is the question, and $a$ is the answer response.
The system uses HyperCLOVA to solve the task by using shots shown in Figure \ref{fig:response-gen-qa}.

\begin{figure}
    \centering
    \begin{minted}[breaklines, fontsize=\scriptsize, frame=single, breaksymbol=\quad, breaksymbolindentnchars=2]{tex}
A friendly Shoko at Odaiba answers the questions of the customers who can't decide which to go between two tourist attractions. Using search information, Shoko gives accurate information.  % Task Description
https://www.overleaf.com/project/630ddcea0207251d52f3c1c5
###

[Search information for Tokyo Tower]
A 333-meter-high radio tower is located in Minato-ku, Tokyo. The name Tokyo Tower was chosen from a public competition. ...  % Summary s1
Ticket prices->Main Deck: Adults 1,200 yen, High School Students 1,000 yen, Children 700 yen, Infants 500 yen  % Selected Data d1

[Tokyo Sky Tree Search Information]
A radio tower located in Sumida-Ku, Tokyo, opened in 2012. With a height of 634 meters, it is the tallest radio tower in the world. ... % Summary s2
Ticket prices->Observation Deck: Adults: 2,300 yen, Medium-aged people: 1,650 yen, Children: 1,000 yen  % Selected Data d2
[Recommended]Tokyo Tower  % Recommended Sight r

Customer: How much is the ticket price for Tokyo Tower?  % Question q
Tourist Information Shoko: According to the search information, a ticket to the main deck of Tokyo Tower is 1,200 yen for adults, 1,000 yen for high school students, 700 yen for children, and 500 yen for infants. It is cheaper than Skytree and is recommended.  % Answer u
    \end{minted}
    \caption{A Shot Example for Response Generation in QA.}
    \label{fig:response-gen-qa}
\end{figure}

\subsection{Closing}
For the closing phase, the robot first informs the customer that the limited time has come, and then moves on to the final persuasion.
During the final persuasion, the robot talks about the experience of visiting the recommended tourist sight.
This is generated by HyperCLOVA.
To generate a plausible experience, the prompt includes user-submitted positive reviews from TripAdvisor.

\subsection{Other Improvements}
In this competition, we are required to use Amazon Polly\footnote{\url{https://aws.amazon.com/polly/}} for text-to-speech conversion.
For some Kanji letters in Japanese, Amazon Polly failed to speak in the right way.
The most critical error in this tourist guidance task was "\begin{CJK}{UTF8}{ipxm}方\end{CJK}", which can be pronounced "kata" or "ho", and it can have different meanings by being pronounced differently.
The right pronunciation can be inferred by reading the context.
Therefore, we used HyperCLOVA to convert "\begin{CJK}{UTF8}{ipxm}方\end{CJK}" to Hiragana letters, so that it could be pronounced correctly with Amazon Polly.

\section{SPEECH RECOGNITION}
\label{sec:speech_recog}
\subsection{Model Architecture}
\begin{figure}[t]
    \centering
    \includegraphics[scale=0.1]{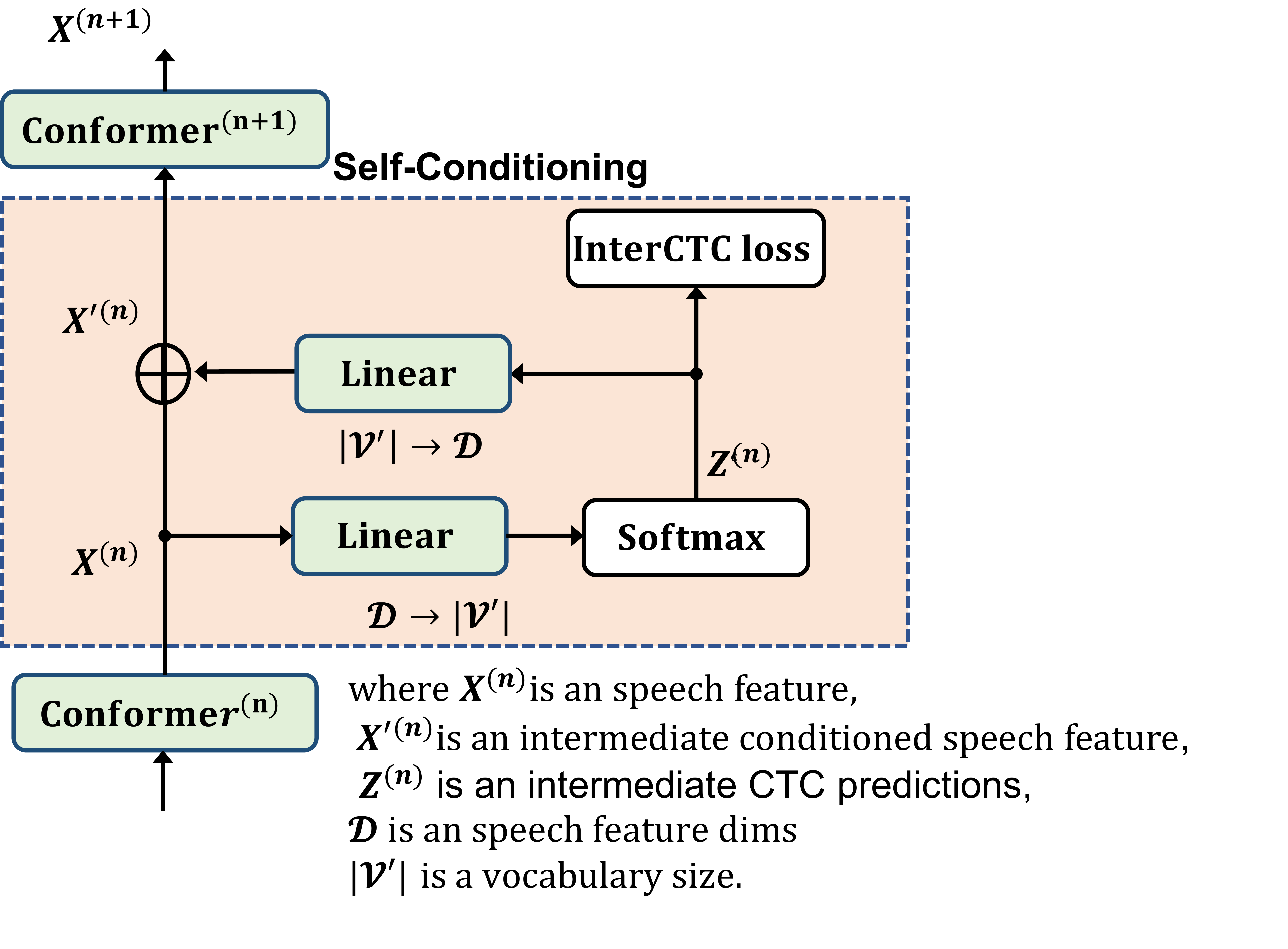}
    \caption{Overview of Self-Conditioning Architecture}
    \label{fig:sc_overview}
\end{figure}

Our original speech recognition system uses an end-to-end model based on connectionist temporal classification (CTC) with self-conditioning architecture \cite{nozaki21_interspeech}.
Our model is composed of stacked encoder layers, which is shown in Figure \ref{fig:sc_overview}.
As shown in the figure, intermediate predictions of the encoder layer are fed back to the next encoder layer.
This self-conditioning architecture is known to improve speech recognition accuracy by relaxing the conditional independence assumption of CTC-based speech recognition models.


%
%
%

\subsection{Training Details}
A following two-stage strategy was adopted for the model training.
Firstly, a general-purpose model was trained using realistic speech data including various domains.
Thousands of hours of in-house speech data were used for the initial training.
Secondly, the model was fine-tuned with another speech data set to match itself to the tourist-guide domain.
This additional training set was created by extracting sentences containing landmarks and place names from the text corpus, which was used for the HyperCLOVA model training.
To reduce the time and effort required for the speech recording, the speech data corresponding to the text sentence was synthesized by our in-house text-to-speech (TTS) engine.




\subsection{Evaluation}
We performed an experimental evaluation of our speech recognition models.
The best-path decoding was employed to reduce latency due to the speech recognition process.
Two evaluation sets were prepared for the general and tourist-guide domains.
The number of utterances was 5,000 for the general domain, and 3,000 for the tourist-guide domain.
TABLE \ref{tab:asr_improvements} compares the character error rates (CERs) between our initial and fine-tuned model.
We can see that the fine-tuned model could improve CER for the tourist-guide domain with keeping that of the general domain.
This means that the model was able to specialize in the tourist-guide domain while maintaining its generality.



\begin{table}[t]
\centering
\caption{Comparison on Character Error Rates (CERs)}
\label{tab:asr_improvements}
\begin{tabular}{|c|c|c|}
\hline
Model      & General domain & Tourist-guide domain \\ \hline
Initial    & 10.90          & 8.37 \\
Fine-tuned & \bf{10.86}     & \bf{3.59} \\ \hline
\end{tabular}
\end{table}

\section{RESULTS AND ANALYSIS}

\begin{figure*}[t]
    \centering
    \includegraphics[width=\linewidth]{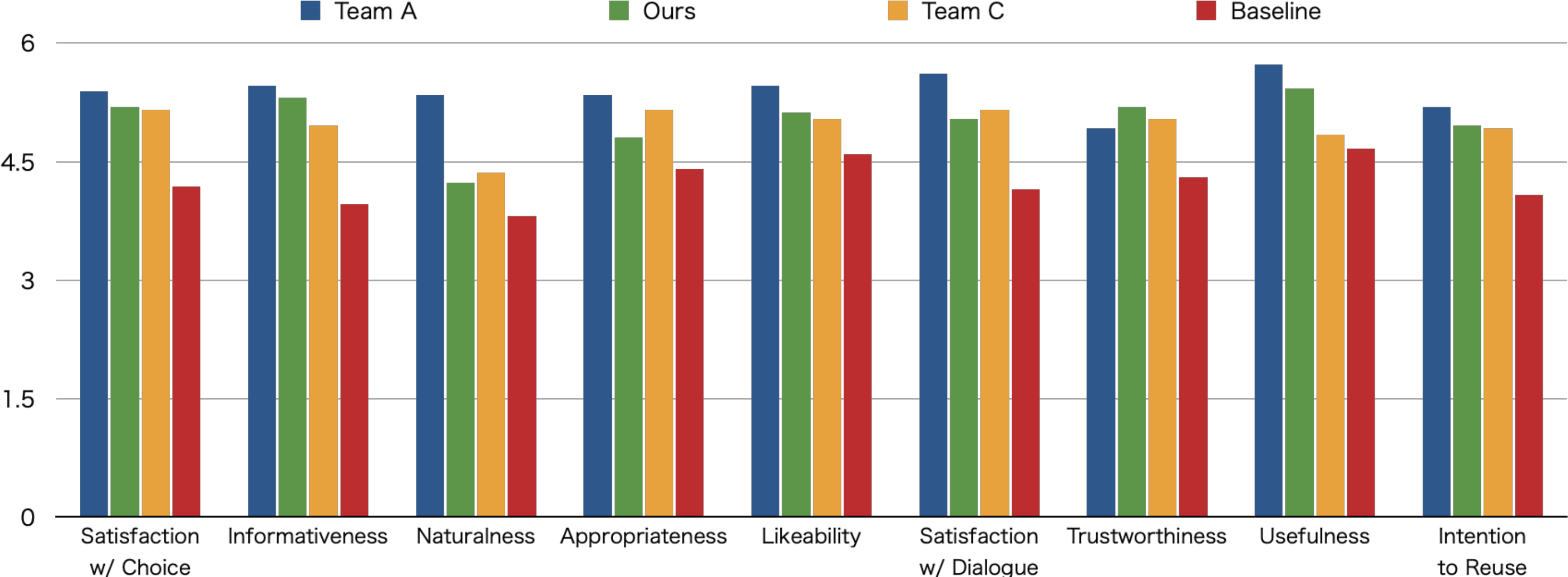}
    \caption{Results of the survey in the Preliminary round. Each item was measured on a 7-point Likert scale.}
    \label{fig:data}
\end{figure*}

Fig. \ref{fig:data} shows the result of the surveys, which was conducted after the dialog session with the robot.
Our system ranked second in overall results in the preliminary round.
\textit{Team A} and \textit{Team C} are the results of first and third place respectively.
\textit{Baseline} system is described in \cite{minato-2022}.

As shown in Fig. \ref{fig:data}, our model has a low score on naturalness.
One possible cause is the generation quality of the HyperCLOVA.
HyperCLOVA often generated unconvincing or out-of-place responses (e.g. "I'm not sure. Please search by yourself.").
The generation quality was degraded mainly because of the lack of contextual information in the prompt and speech recognition errors.
Another possible reason is the response time which takes 2 to 4 seconds, and users could feel stressed waiting for the response.
For the trustworthiness metric, our system was the best among all the submitted systems.
This could be because our model can generate flexible responses that reflect the user's input, which may give impressions to the customers that they are being listened to.

\section{CONCLUSIONS}

This paper explains the tourist guidance robot based on HyperCLOVA, which was submitted to the Dialogue Robot Competition 2022.
Our model has used HyperCLOVA to solve multiple types of language tasks in tourist guidance, including summarization, information extraction, response generation, style transfer, and more.
We also implemented a speech recognition system fine-tuned for this dialog task.
As a result of the qualifying round, our model achieved second place for the overall score and moved on to the final round.
The evaluation results show that our system could help users feel trustworthy, while leaving huge space for improvements on naturalness.

\addtolength{\textheight}{-10.6cm}   

\bibliographystyle{IEEEtran}
\bibliography{root}

\end{document}